\def\year{2021}\relax
\documentclass[letterpaper]{article} 
\usepackage{aaai21}  
\usepackage{times}  
\usepackage{helvet} 
\usepackage{courier}  
\usepackage[hyphens]{url}  
\usepackage{graphicx} 
\usepackage{natbib}  
\usepackage{caption} 
\usepackage{mathtools}
\usepackage{amsmath}
\usepackage[switch]{lineno}
\DeclarePairedDelimiter\floor{\lfloor}{\rfloor}
\newcommand*\mean[1]{\bar{#1}}
\nocopyright  
\usepackage{makecell}
\usepackage{tabularx}

\title{Brain2Word: Decoding Brain Activity for Language Generation}
\author{
    Nicolas Affolter, 
    B\'eni Egressy, 
    Dami\'an Pascual, 
    Roger Wattenhofer\thanks{Authors in alphabetic order.}
    \\
}
\affiliations{

    ETH Zurich\\

    nicolaff@student.ethz.ch, \{begressy,dpascual,wattenhofer\}@ethz.ch

}

\begin{document}

\maketitle



\begin{abstract}
Brain decoding, understood as the process of mapping brain activities to the stimuli that generated them, has been an active research area in the last years. In the case of language stimuli, recent studies have shown that it is possible to decode fMRI scans into an embedding of the word a subject is reading. However, such word embeddings are designed for natural language processing tasks rather than for brain decoding. Therefore, they limit our ability to recover the precise stimulus. In this work, we propose to directly classify an fMRI scan, mapping it to the corresponding word within a fixed vocabulary. Unlike existing work, we evaluate on scans from previously unseen subjects. We argue that this is a more realistic setup and we present a model that can decode fMRI data from unseen subjects. Our model achieves $5.22\%$ Top-1 and $13.59\%$ Top-5 accuracy in this challenging task, significantly outperforming all the considered competitive baselines. Furthermore, we use the decoded words to guide language generation with the GPT-2 model. This way, we advance the quest for a system that translates brain activities into coherent text.
\end{abstract}


\section{Introduction}

Recent advances in brain imaging suggest that it may be possible to infer what a person is perceiving from their brain scans. The ability of decoding brain signals has important applications in medicine, e.g., assisting handicapped people who cannot move or talk, as well as in the consumer industry, e.g., producing content that adapts to what a person is seeing, feeling or thinking. In this context, language is of particular interest, since it is the vehicle we use to express our thoughts.
A body of research has focused on decoding functional Magnetic Resonance Imaging (fMRI) scans into a representation of the word a person is reading while being scanned. By measuring similarity between the decoded scan and actual word representations, they show that the decoded representation is closer to its corresponding word representation than to another word with a chance significantly higher than random. 

Although an important first step in showing that inferring such information from brain scans is at all possible, this task is rather simple and has limited potential applications. The inference models used to solve it are equally simple, normally based on ridge regression or simple Multi-Layer Perceptrons (MLP), while they rely on complex subject-specific pre-processing and feature selection \cite{pereira2018toward, sun2019towards}. 
In this work, we argue that a more demanding setup needs to be considered in order to understand the extent to which we can currently map brain activities to words. In particular, we propose direct classification, i.e. to directly classify a brain scan as one of the $v$ words within the considered vocabulary, as opposed to pairwise classification. Furthermore, we address brain decoding on unseen subjects, i.e., the training data does not contain any data from the test subject. This is known to be a remarkably hard problem, since fMRI scans are very different across subjects and even across recording sessions, among other reasons due to variable numbers of voxels and lack of alignment between scans. 
Thus, the challenge with this setup is twofold, the evaluation task is more demanding and strong generalization is required since subject-specific pre-processing is not possible.

On the bright side, in this setup we can exploit a larger training set consisting of the scans from $n-1$ subjects in order to train more complex models, where $n$ is the number of subjects in the dataset. We propose a neural autoencoder model that takes as input a complete fMRI scan and outputs the stimulus word. We use minimal external knowledge, specifically the Regions of Interest (ROIs) of the brain scan, and let the model learn features that generalize to all subjects. We validate our model on the classical pairwise classification task and then, we demonstrate its performance in direct classification.

Then, we take a novel research direction 
and consider a practical application of brain decoding. We envision a system that decodes concepts from the brain, rather than complete sentences, and uses these concepts to guide the generation of coherent text. Such a system could help individuals with speech impairments to communicate. 
To this end, we leverage GPT-2, a recently proposed model for language generation which can produce outstandingly realistic text. We condition GPT-2 with the decoded brain scans and show, as a proof-of-concept, that brain activities can guide language generation. Although a long path still needs to be covered before having a fully functional system, our work sets a first stone towards translating brain activities into coherent text. 

All in all, our contributions are\footnote{Code: \url{https://github.com/nicolaffETHZ/Brain2Word_paper}}:
\begin{itemize}
    \item We propose a new and more challenging evaluation setup for brain decoding, i.e., to decode the brain activation from a subject unseen during training directly into a specific word in a bounded vocabulary.
    \item We present a neural network-based model that improves fMRI-to-word decoding by a significant margin in the existing evaluation framework as well as in our more challenging and realistic setup.
    \item We bridge fMRI decoding to a real-world application: language generation conditioned on brain activities.
\end{itemize}

\section{Related Work}

\subsection{Brain Decoding}

Since the publication of the seminal work \cite{mitchell2008predicting}, decoding brain activity into words has attracted a lot of attention from the research community. In recent years, a large number of studies has tackled this problem from different angles. \cite{palatucci2009zero} proposed a model to learn new classes unseen during training, \cite{just2010neurosemantic, huth2016natural, handjaras2016concepts} built brain decoders that helped them draw conclusions about the way the brain processes language. \cite{pereira2018toward} presented a model that decodes brain activity into word embeddings. \cite{wehbe2014simultaneously} decoded text passages rather than single words and, similarly, \cite{sun2019towards} decoded sentences using distributed representations. These works represent just a part of a large body of research \cite{wang2020probing,schwartz2019inducing,kivisaari2019reconstructing} that has strongly contributed to the progress of decoding and understanding brain activities. 

In most existing literature the scans used for training come from the same subject that is evaluated. Due to the misalignment of brain scans between subjects, evaluating on a different subject is a very challenging problem. Recent work \cite{van2018modeling, nastase2020leveraging} has studied this problem in controlled settings and approached it from an algorithmic perspective. Here, we take a data-driven approach to successfully generalize brain decoding to unseen subjects.






\subsection{Language models and the brain}

The field of Natural Language Processing (NLP) has undergone outstanding progress in the last few years thanks to a family of deep learning models called Transformers \cite{vaswani2017attention, liu2019roberta, raffel2019exploring}. These models are currently state-of-the-art in most NLP tasks and remarkably, in language generation, e.g., the GPT-2 model \cite{radford2019language}.
Recent work has tried to establish a link between the brain and these models. \cite{gauthier2019linking} decodes fMRI to improve latent representations inside a transformer for NLP tasks. \cite{toneva2019interpreting} use fMRI scans to interpret and improve BERT \cite{devlin2018bert}, a well-known transformer. Relatedly, \cite{muttenthaler2020human} use EEG features to modify attention weights in an LSTM based model.

Different from prior work, we devise a direct application of brain decoding: to use brain activities in order to guide language generation with GPT-2. \cite{nishimoto2011reconstructing} demonstrated that it is possible to dynamically decode brain activity in the form of fMRI scans. Based on this result, we advance towards a brain-computer interface capable of translating brain activity into coherent text.




\section{Background}

We call \emph{brain decoder} or simply \emph{decoder} a function capable of mapping brain activities to the stimulus that generated them. 
In this work, we map brain activities in the form of fMRI scans to the text presented to subjects during scanning. 
We consider two types of decoders, first, classical regression-based decoders \cite{bulat2017speaking} which learn a parametric mapping from the fMRI scan to a vector representation of the text; and second, we propose classification-based decoders, which learn to map brain activities to a word within a bounded vocabulary. 

\subsection{Dataset}

We use the dataset from \cite{pereira2018toward}. This dataset contains fMRI scans from $15$ subjects. Each subject was recorded reading $180$ different words, one at a time. Each word, was shown to the subject following three different paradigms that ensure that all subjects focus on the same meaning, i.e., supporting the word with a word cloud, with sentences and with images. Additionally, $8$ of the subjects were scanned while reading sentences from a dataset that consists of $384$ sentences from $96$ different passages. Finally, $6$ of the subjects (with overlap with the $8$ previous subjects) were also scanned reading another dataset of $243$ sentences from $72$ passages. $6$ subjects were not recorded reading sentences.\footnote{For more details on the dataset refer to \url{https://osf.io/crwz7}}

In this work, we are interested in decoding individual words and so, our dataset consists of $15$ subjects with $540$ scans each ($180$ words, three paradigms). However, we also explore pretraining our model with the sentence recordings.

\subsection{Evaluation Tasks}


\paragraph{Pairwise classification} In this task, a regression-based decoder is trained to produce a vector representation from a brain image (fMRI). 
Then, for each possible pair of words the correlation between the decoded vectors and the actual embedding vectors of both words is computed, i.e., four values. If the decoded vectors are more similar to their corresponding word embeddings than to the alternatives, the evaluation is considered correct. As such, the random baseline for this task is $0.5$. The final result is the mean across test instances.

This task presents certain limitations arising from the representation the brain image is decoded into. In \cite{pereira2018toward}, this representation is a GloVE embedding \cite{pennington2014glove} and therefore, it contains information beyond semantics, such as word frequency in the data used to train the embedding. \cite{gauthier2018does} show that decoding brain images into representations derived from models optimized to solve very different tasks, e.g., image captioning or machine translation, produce similar results as the baseline decoder from \cite{pereira2018toward}. This suggests that training the decoder to produce a certain representation vector is fundamentally limited by the type of representation used. Therefore, as an additional evaluation task, we propose direct classification.

\paragraph{Direct classification} In this task, a classification-based decoder receives as input a brain scan and produces as output a vector of size $v$, where $v$ is the size of the vocabulary. In our case $v=180$. This vector contains the predicted probability for each word in the vocabulary. This way, the decoder is effectively a classifier that infers which word is seen by the subject when the scan is taken. This task is significantly more challenging than the pairwise classification task, with the random baseline being $1/v$ for the Top-1 score ($0.06\%$ in our case). On the other hand, it does not suffer from limitations associated with the chosen vector representation. In this task, we report Top-1 and Top-5 scores, i.e., the classification is correct if the word is within the Top-X predictions.

\subsection{Evaluation Setup}

Most previous work on brain decoding \cite{pereira2018toward, sun2019towards} has considered the scenario where the model is trained with data from the same subject that is being evaluated. We argue that this scenario is not suitable for real life applications and that it in fact limits our ability to decode brain activities. 

In said scenario, for each new subject a training set needs to be recorded in order to train a personalised decoder model. Recording fMRI scans is a costly and slow process, e.g., for one subject in our target dataset it takes approximately $4$ hours just to obtain the $180$ brain scans, which still have to be processed.\footnote{Five repetitions per paradigm (three paradigms), where each repetition needs two runs ($90$ words per run) and each run takes 8 minutes \cite{pereira2018toward}.} Therefore, it is desirable to have models that are able to decode the brain activities of new subjects without the need for subject-specific training data.
Furthermore, the amount of data that can be recorded for one subject is limited, which restricts the complexity of the decoders and forces the model to rely on subject-specific pre-processing. By using the data from all recorded subjects, we can build larger neural network-based decoders that learn general features across subjects, dispensing with the need for subject-specific processing steps.
This however is a difficult problem since there is almost no alignment between the fMRI scans of different subjects.

In this work, we consider this challenging setup and follow a leave-one-out strategy in our evaluation. This way, we train our model with the data from $n-1$ subjects and test it on the remaining subject; we repeat this process for each subject. This simulates a real-world scenario where an existing model is applied to a new, as yet unseen, subject.

\section{Brain Decoding Model}

\begin{figure}[t]
\centering
\includegraphics[width=\linewidth]{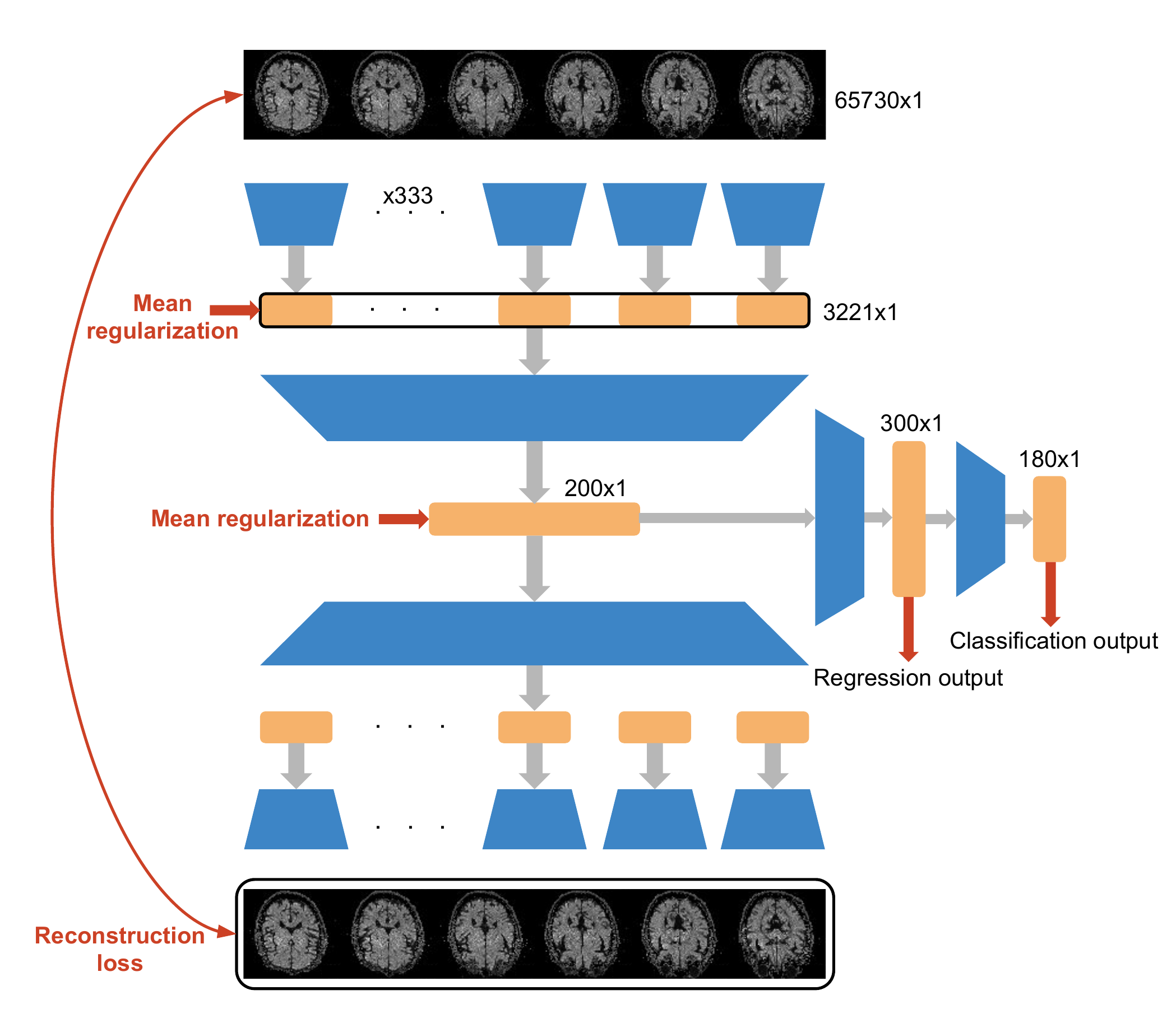}
\caption{Architecture of the improved decoder. The blue trapezoids represent dense layers, the orange rectangles feature maps and the solid black lines concatenation. The shape of the feature maps is specified, as well as the points where the different regularization terms are applied.}
\label{fig:arch}
\end{figure}

We propose a new model that leverages recent advances in deep learning in order to decode brain activities in the form of fMRI scans into text. Our model can be implemented as either a regression-based or a classification-based decoder by simply changing the last layer and the related term of the loss function. The regression-based decoder has a final linear layer that outputs a vector of the size of the target representation; following \cite{pereira2018toward} we use GloVE embeddings of size $300\times1$. The regression loss is calculated on this output and has the form: 

\begin{equation*}
    \begin{aligned}
    \mathcal{L}_{reg} = \sum_{i}^{v} \left( \cos \left(  y_{pr,i},y_{true,i} \right) 
     - \sum_{j \neq i}^{v} \cos \left(y_{pr,i},y_{true,j} \right) \right)
    \end{aligned}
\end{equation*}

Where $y_{pr,i}$ is the predicted word embedding for word $i$, $y_{true,j}$ is the real word embedding for word $j$ and $\cos(x,y)$ is the cosine distance between vectors $x$ and $y$. 
Note that to ease notation this formulation corresponds only to one paradigm of one subject, the total loss is calculated by summing over all paradigms for each subject in the training set. The same applies to the formulation of the other loss terms presented below.
This loss is inspired by the triplet loss \cite{schroff2015facenet} and aims at guiding the model's output as close as possible to the true embedding while keeping it as far as possible from the embeddings of the other words in the vocabulary. We observed that this loss function helped to prevent the model from collapsing towards a mean representation of the word embeddings.

In the classification based decoder, the regression layer of size $300\times1$ is turned into a non-linear layer and an additional softmax layer is added on top of it. This way the model outputs a one dimensional vector of probabilities of the size of the vocabulary $v$, $180\times1$ in our case. The classification loss is given by the categorical cross-entropy between the vector of predicted probabilities $y_{pr}$ and the one hot representation of the target word $y_{true}$:

\begin{equation*}
    \mathcal{L}_{class} = - \sum_{i}^{v} y_{true,i} \cdot log(y_{pr,i})
\end{equation*}

Where $i$ is the index for a given word in the vocabulary.

We first consider a simple model which takes as input a one dimensional fMRI scan of size $65,730\times1$ voxels and generates a latent vector of size $200\times1$; the input size is such that all the scans in the dataset fit, if a scan is smaller we pad it. The latent vector is used to produce either the regression or the classification target. Apart from the regression and classification layers, the model consists of two non-linear fully connected layers that produce feature maps of size $2000\times1$ and $200\times1$ respectively. Each non-linear layer has $0.4$ dropout, batch normalization and Leaky ReLU activation ($\alpha = 0.3$). We take this simple model as a base model and improve it with the extensions detailed below. We do an ablation study of the extensions in the results section to measure their impact on the performance of the model. The complete model is depicted in Figure~\ref{fig:arch}.

\paragraph{Regions of Interest} Some areas of the fMRI scans can be identified as Regions Of Interest (ROIs) following the brain atlas from \cite{gordon2016generation}, provided with the dataset. This atlas defines areas of gray matter in the brain responsible for different brain functions, in particular for perception and language. 
Each ROI is associated with one or more brain functions. 
To exploit this knowledge and reduce the size of the model, we restrict ourselves exclusively to these ROI regions and process each of them separately in the first layer of our model. This way, we use one dense layer for each of the $333$ ROIs from the atlas and concatenate their outputs. Note that the ROIs vary in size, and thus, so do the individual dense layers. We set each of these to produce an output vector of size $max(\floor{\frac{ROI Size}{20}},1)$, where the factor $20$ is a hyperparameter chosen to regulate the size of the hidden layer. On our hyperparameter search we found this value to be adequate.

\paragraph{Autoencoder} We turn the model into an autoencoder (decoder-encoder) by adding an encoder that mirrors the base model, i.e., the decoder. This encoder reconstructs the input brain activities (fMRI) from the latent vector and to this end, we add a reconstruction term to the loss function. The reconstruction loss is given by:

\begin{equation*}
    \mathcal{L}_{rec} = \sum_{i}^{v}\cos \left(  x_{out,i},x_{in,i} \right)
\end{equation*}

Where $x_{in}$ is the input fMRI scan and $x_{out}$ is the reconstructed fMRI, i.e., the output of the encoder. The rationale behind using an autoencoder is that the reconstruction loss should help learning by increasing the training signal and by acting as a regularizer.

\paragraph{Mean regularization} Since the model should produce the same output for scans from different subjects when exposed to the same word, the latent representations inside the model should converge. In other words, we expect the model to progressively discard subject-specific physiological information in order to extract the word the subject is reading. To this end, we regularize the output of each layer of the decoder to be similar to the mean representation for a given word across subjects at that layer and dissimilar to the mean representation of the other words. More formally, we add a term to the loss function with the same structure as the regression loss:

\begin{equation*}
    \begin{aligned}
    \mathcal{L}_{mean} =  \sum_{i}^{v} \left( \cos \left( h_{i}^{(l)},\mean h_{i}^{(l)}  \right) - \sum_{j \neq i}^{v} \cos \left( h_{i}^{(l)},\mean h_{j}^{(l)} \right) \right) 
    \end{aligned}
\end{equation*}

Where $h_{i}^{(l)}$ is the predicted hidden representation of word $i$ in layer $l$ and $\mean h_{i}^{(l)}$ is the mean of the predicted hidden representations for word $i$ at layer $l$ across all subjects; these mean representations are updated after every epoch. At the beginning of the training the model produces meaningless representations and, for this reason, we first train without mean regularization and only when learning saturates, the mean regularization is activated and the model resumes training until early stopping occurs. 

\paragraph{Unsupervised pretraining} As detailed above, the dataset from \cite{pereira2018toward} contains two additional sets of fMRI scans, amounting to a total of $4,530$ scans. These scans were recorded while subjects read sentences, instead of the words that conform our learning target. Therefore, we can only use these additional scans in an unsupervised manner. The autoencoder structure of our improved model allows us to do this: we pretrain our model on the sentence scans using exclusively the reconstruction loss $\mathcal{L}_{rec}$ for $30$ epochs. Afterwards, we start the supervised training phase on the word scans. With the pretraining phase we aim to exploit general language-related fMRI features in order to place our model at a better starting point. This can help the model to eventually reach a better minimum on our training objective.


\section{Results}

As explained above, we follow in our experiments the leave-one-out approach, i.e., we use all subjects except one for training and we evaluate on this left-out subject. This process is repeated for all subjects. We use subject $M15$ for validation and the rest for testing. We perform the ablation study on the validation subject. Likewise, the hyperparameters used in our model and described so far are also found by grid search evaluated on the validation subject. The evaluation on the remaining $14$ test subjects is used for the final results on the pairwise and classification tasks.

As already mentioned, for the calculation of the pairwise accuracy, we follow \cite{pereira2018toward} in all cases and use GloVE embeddings as the decoding target.

\subsection{Ablation Study}

In this study, we first consider the base model and then progressively add the extensions in the order they were presented. We evaluate both versions of the model: the regression-based and classification-based decoder. In Table~\ref{tab:ablation}, we report the pairwise classification accuracy for the regression-based decoder, and the Top-1 and Top-5 scores for the classification-based decoder.

\renewcommand{\arraystretch}{1.3}
\begin{table}[h]
\centering
\begin{tabular}{l||c||c|c}
Model & Pairwise & Top-1 & Top-5 \\
\hline
Base & 0.8268 & 4.07\% & 11.66\% \\
\hline
+ ROI & 0.8336 & 4.25\% & 11.85\% \\
+ Reconstruction & 0.8411 & 4.81\% & 12.96\% \\
+ Mean reg. & 0.8464 & 5.55\% & 13.14\% \\
+ Pretraining & \textbf{0.8637} & \textbf{6.29\%} & \textbf{15.00\%} \\
\end{tabular}
\caption{Ablation study. The extensions are progressively added to the model.}
\label{tab:ablation}
\end{table}
\renewcommand{\arraystretch}{1}

We see that all four extensions monotonically improve the performance of the model for the three metrics. This study validates our design choices and thus, in the remaining experiments we use the complete model.

Finally, we want to further investigate the hypothesis that mapping brain activities to word embeddings is limiting the model by introducing unwanted information in the decoding target (e.g., word frequency) \cite{gauthier2018does}. To this end, we add to our best classification model the regression loss as an additional optimization objective. If the word embedding was a good representation of concepts, this additional term would help (by increasing the training signal), or at least not harm the classification performance. However, adding this loss term to the complete classification model degrades the Top-1 and Top-5 scores down to $5.89\%$ and $13.55\%$ respectively. This supports the hypothesis that the GloVE embedding is a noisy representation of the concept, which further underscores the need for a better evaluation task, such as our proposed direct classification.

\subsection{Pairwise Classification}

To put our model into context with respect to existing work, we evaluate it on the pairwise classification task and compare it with four competitive baselines. First, we take the model from \cite{pereira2018toward} which uses ridge regression, in the following we refer to this model as Universal Decoder. Second, we take a simple MLP consisting of a non-linear layer that maps the input to a feature vector of size $2000\times1$ followed by a linear layer that outputs the regression target, i.e., a GloVe embedding ($300\times1$). Third, we take a big MLP with one non-linear dense layer per ROI, as in our complete model, followed by a linear layer that outputs the GloVe embedding. Last, we evaluate the VQ-VAE model from \cite{van2017neural} adapted to regression-based decoding of fMRI. This model discretizes the latent space, thus, we hypothesize that it may naturally separate the scans according to the word that they encode. 

\begin{figure}[t]
\centering
\includegraphics[width=\linewidth]{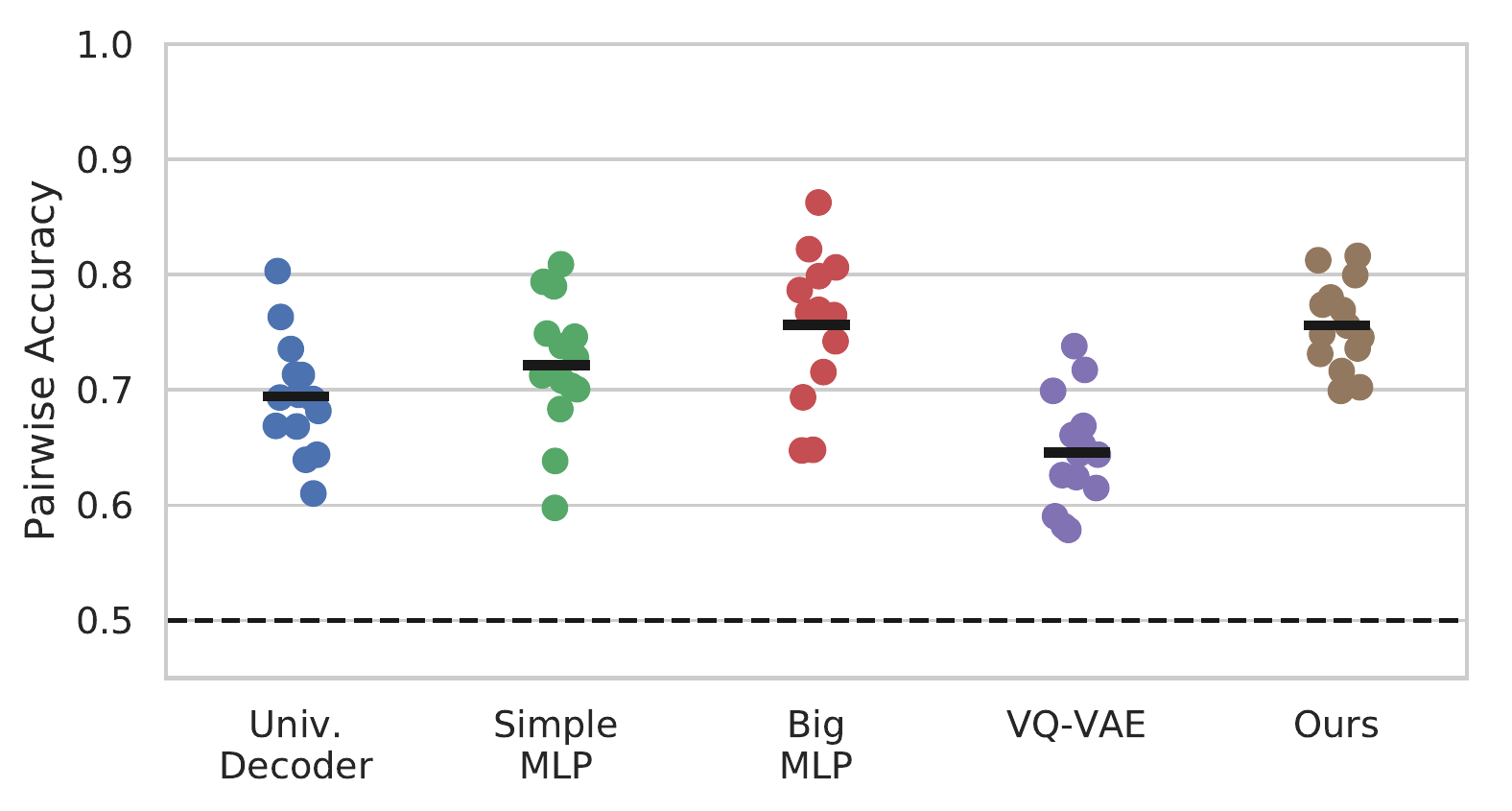}
\caption{Pairwise accuracy of the different models. Each point represents a subject, the solid lines are the mean across subjects and the dashed line the random baseline.}
\label{pair_results}
\end{figure}

Given the reduced capacity of the Universal Decoder, training it on subjects different than the test subject produced close to random performance. Therefore, we train and evaluate it in the same manner as in the original work \cite{pereira2018toward}, i.e., for each paradigm $170$ words of a given subject are used for training and the remaining $10$ for testing. This is repeated $18$ times over to cover all the words and the results are averaged. This is different from the other models where we follow the leave-on-out approach, i.e., the target subject is never seen during training.

We report the results on all the $14$ test subjects in Figure \ref{pair_results}. 
First, we see that the VQ-VAE has the worst performance, which rejects our hypothesis about the discrete latent space. All the other models outperform the Universal Decoder even with the disadvantageous training setup (unseen test subject). It is also noteworthy that the ``Big MLP" performs on par with our model in this task, albeit with higher variance across subjects. These results show that neural network-based decoders successfully generalize to unseen subjects and even clearly outperform classical models trained on the target subject.

\subsection{Direct Classification}

Next, we evaluate our model in the direct classification task. We compare our model against five competitive baselines, the same four as above adapted to the classification task, and additionally, against Principal Component Analysis decomposition (PCA), for dimensionality reduction, followed by XGBoost \cite{chen2016xgboost}, a tree-based classification algorithm. To perform classification using the Universal Decoder we take the output of the model and do nearest neighbour search on the GloVE embeddings of the $180$ words of our vocabulary. For the other models, we turn the last layer into a classification layer by adding softmax and changing the output size to $180$, i.e., the number of classes.

\begin{figure}[t]
\centering
\includegraphics[width=\linewidth]{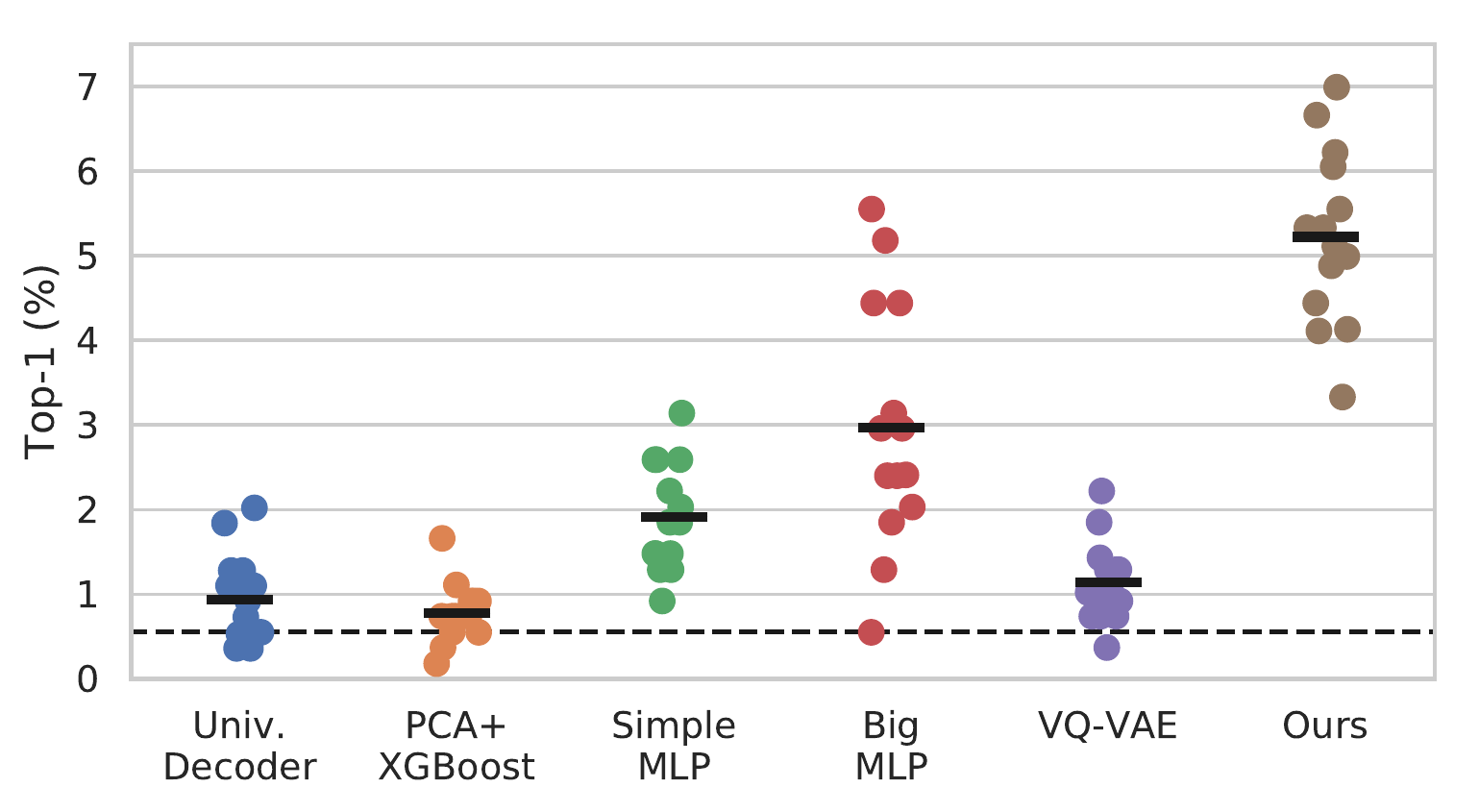}
\caption{Top-1 Score across models.}
\label{top1}
\end{figure}

\begin{figure}[t]
\centering
\includegraphics[width=\linewidth]{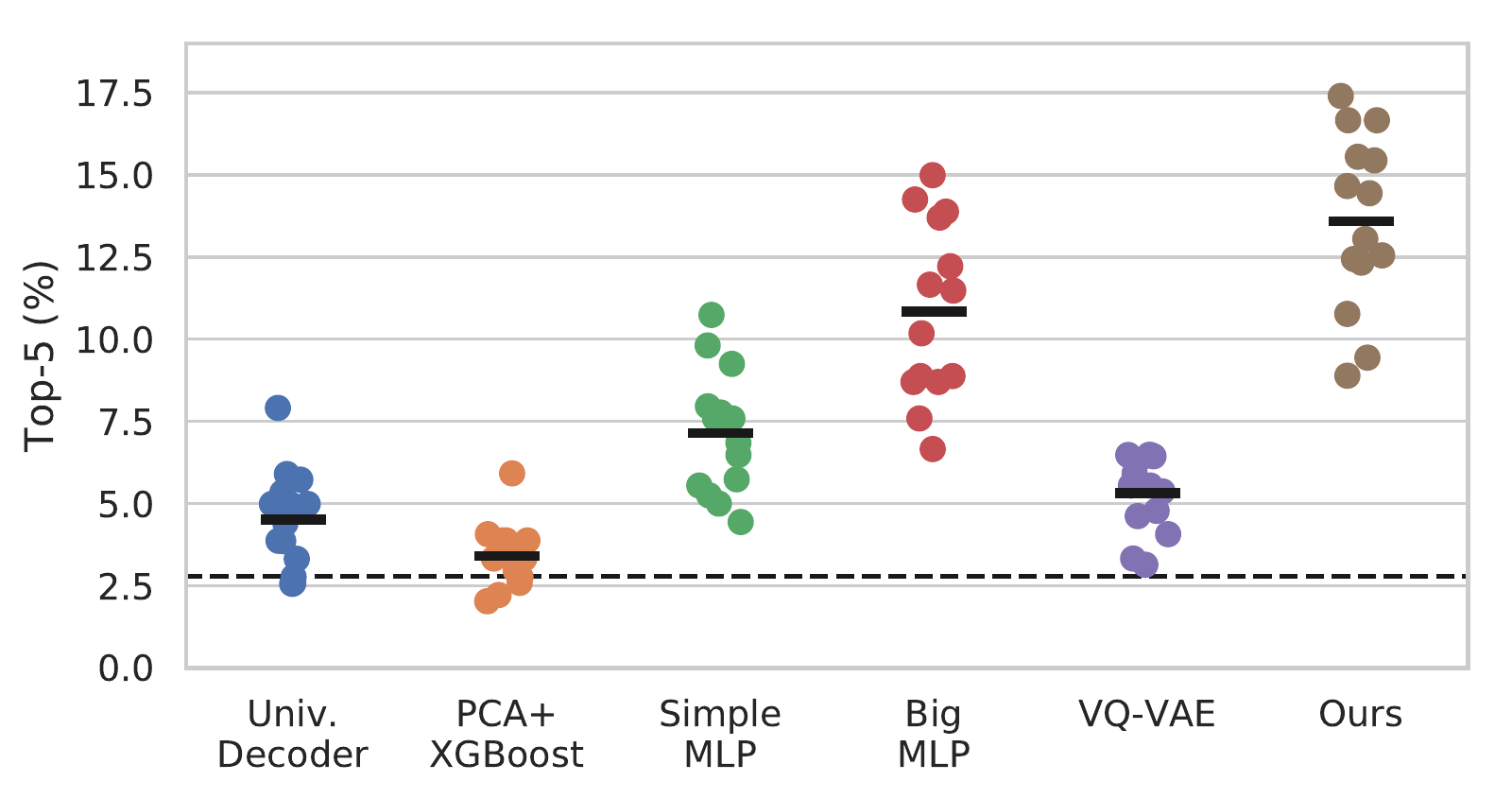}
\caption{Top-5 Score across models.}
\label{top5}
\end{figure}

We represent the results in the same manner as in the previous section, Figure \ref{top1} shows the Top-1 scores and Figure \ref{top5} the Top-5. In this more complicated task (the random baseline is $0.6\%$ for Top-1 and $2.8\%$ for Top-5) the Universal Decoder mean accuracy is $0.94\%$ for the Top-1 score and $4.5\%$ for Top-5, slightly above random. Again, the VQ-VAE performs the worst among the neural models with scores similar to those of the Universal Decoder, and PCA plus XGBoost performs very close to random. On the other hand, the Top-1 mean accuracy for the simple MLP, big MLP and our model is $1.91\%$, $2.97\%$ and $5.22\%$ respectively. We see that in this challenging setup, our complete model is clearly the best for both Top-1 and Top-5 scores. In particular, its Top-5 mean accuracy is above $13.59\%$, almost 5 times the random baseline. This result is outstanding given the difficulty of the task, i.e., decoding the exact word corresponding to the fMRI scan of an unseen subject.

The good performance of our decoder on this realistic scenario shows the potential of using brain decoding in real life applications. Next, we show a proof-of-concept of how language generation can be guided by decoded fMRI scans.

\section{Bridging fMRI to Language Generation}


\begin{figure}[t]
    \centering
    \includegraphics[width=\linewidth]{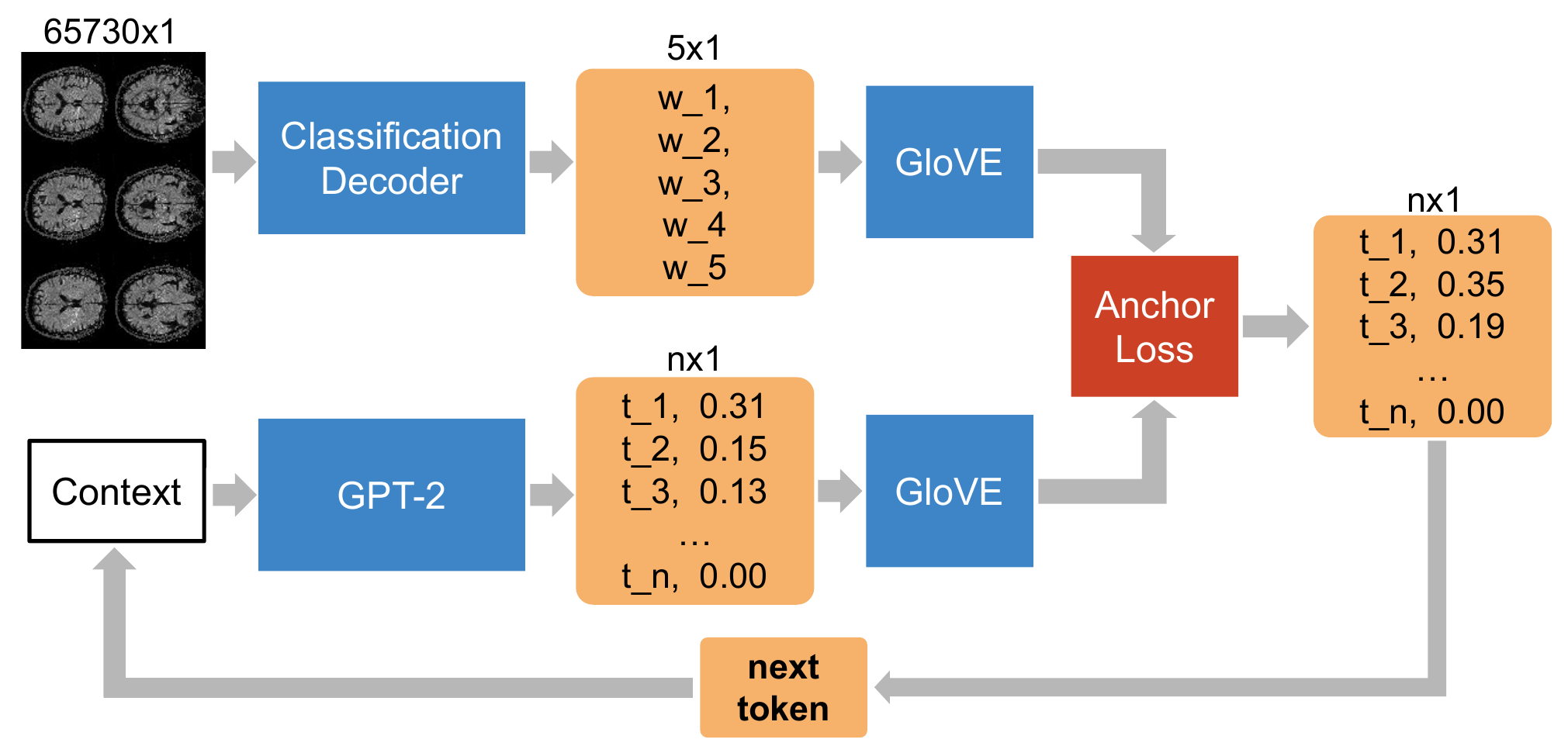}
    \caption{Architecture of the conditioned language generation model. Our classification decoder finds anchor points for the language generation model (GPT-2), which generates the next token given some initial context. }
    \label{fig:fMRI_GPT2}
\end{figure}

Next, we present an approach for combining information from the fMRI scans with a powerful language model. In particular, we generate text conditioned on the fMRI data using GPT-2 \cite{radford2019language}.

On the one hand, we use our fMRI classification decoder to transform a brain scan into a probability vector over our vocabulary of $180$ words. We select the top $5$ predictions, $w_1, w_2, ..., w_5$, and calculate their GloVE embeddings. We will use these embeddings as anchor points for the language generation model.
On the other hand, the GPT-2 model autoregressively generates text, i.e., from previously generated context. To generate a new word, it processes the past context and produces a vector of probabilities over the whole vocabulary, then it samples the next word from the top-$k$ words with the highest probabilities. We denote this vector of probabilities by $\Vec{p} = \{p_1, p_2, ..., p_m \}$, where $p_i$ is the probability corresponding to token $t_i$ from the GPT-2 vocabulary and $m=50257$ is the size of the GPT-2 vocabulary. 

To guide language generation, we modify this vector of probabilities. We adjust the next token prediction scores $\Vec{p}$ based on the cosine distance in the GloVE space between each word in the GPT-2 vocabulary and the anchor points. We use the GloVE embeddings to give a common space where the words decoded from the fMRI scan and the GPT-2 token predictions can be compared. 
This additional term guides the next token generation towards the anchor points. The adjusted scores $\Vec{p'}$ are calculated as

$$p'_i = p_i + k \sum_{j=1}^{5} \cos \left( \gamma(t_i), \gamma(w_j) \right)$$

Where $t_i$ is the i\textsuperscript{th} token of the vocabulary, $\gamma$ denotes the GloVE embedding and $k$ is a hyperparameter controlling how heavily the scores are influenced by the anchors $w$. 
Finally, GPT-2 samples the next token from the top-$k$ tokens with the highest updated probability score. The process repeats with the new token added to the context. Our complete model is illustrated in Figure \ref{fig:fMRI_GPT2}. 

We emphasize that this is a general approach for conditioning language generation models on external input. Anchor points from any upstream model can be used to steer language generation. Moreover the generative model can also be replaced. Our model uses GloVE embeddings to connect the two parts together, but again any other word embedding scheme could be adopted.


\subsection{Results}

Now, we evaluate the end-to-end application of our fMRI language generation model. We take fMRI images as the input to the brain decoder. In this proof-of-concept 
we restrict our experiments to the fMRI scans where the classification decoder has a correct Top-5 prediction, that is the correct word appears in the top $5$ words. We use a dataset of $40$ brain scans for test and $10$ for validation, which we use to tune the value of the anchor term~$k$; we settle on $k=7.0$.
For the language generation we use the GPT-2 \emph{small} model inputting snippets from the Harry Potter books \cite{HarryPotterBooks2018} as initial context. The snippets are made of $2$ consecutive sentences randomly selected and to avoid topic-specific content we filter out snippets with proper nouns. 

First, we present an objective evaluation of the fMRI-conditioned language generation model. We use perplexity to quantify the fluency of the text and relevant word count as a measure of the success of the conditioning. The perplexity is calculated based on the direct output of the GPT-2 model, before the anchoring term is added. This ensures that our evaluation of perplexity is not biased due to the increase of the token probabilities through the anchor term.
The word count represents how well the generation is guided towards the semantic content of the fMRI scan. We count both, the number of occurrences of the correct word corresponding to the fMRI scan, as well as the number of occurrences of the $10$ nearest neighbours in the GloVE space.
We perform $10$ runs per brain scan with the same random $10$ initial contexts for each fMRI scan. In each run we generate $30$ tokens, which is approximately 1 or 2 sentences. As a comparison we take the vanilla GPT-2 predictions with no conditioning on the fMRI data and evaluate the same metrics. 

\renewcommand{\arraystretch}{1.3}
\begin{table}[h]
\centering
\begin{tabular}{l||c|c|c}
Model & Perplexity & \makecell{Word Count} & \makecell{Rel. Words} \\
\hline
GPT-2 & 89.24 & 0.00 & 0.04 \\
\makecell[vl]{+ anchoring} & \textbf{50.64} & \textbf{0.52} & \textbf{1.02} \\
\end{tabular}
\caption{Comparison of language generation with and without conditioning on fMRI brain scans.}
\label{tab:gpt2_vs_anchoring_obj}
\end{table}
\renewcommand{\arraystretch}{1}

The results can be seen in Table \ref{tab:gpt2_vs_anchoring_obj}. As expected, the vanilla GPT-2 predictions have a $0.00$ average word count and an average related words count $0.04$: 
with a vocabulary size of $m=50,257$ it is highly unlikely that the generated text contains the desired word and so, this serves to set the random baseline. With anchoring the average word count increases to $0.52$ and the related words to $1.02$, which is significant since we only generate $30$ words, and demonstrates the success of our guided generation approach. Moreover the fluency of the generated texts does not appear to suffer, in fact the average perplexity score improves compared with the benchmark. We hypothesize that anchoring helps the generation to revolve around a reduced set of topics, reducing the chances of generating low probability words.

For a qualitative analysis, we present an example of text generated by our model in Table \ref{tab:gpt2_vs_anchoring}. To study how the anchors affect the generation, we compare it to text generated without anchoring.
We see that the model produces coherent text and that the target word does appear. However, no punctuation tokens are generated and although this is not a big issue given the length of the text, it shows a direction to improve our conditioning strategy. Also, note that bigger generation models would improve the quality of the generated text.

\renewcommand{\arraystretch}{1.3}
\begin{table}[ht]
\centering
\begin{tabularx}{\columnwidth}{X}
Context: \\
\textit{``Everyone stand by a broomstick.  Come on, hurry up."} \\
\hline
Anchor: \\
\textbf{level}, picture, sign, mechanism, device \\
\hline
With anchoring:\\ 
\textit{If the team is not up to the same \textbf{level} as the other team then they will have a hard time even if they work up a number of good}
\\
\hline
Without anchoring:\\ 
\textit{"That's a DMT / If you can do that, your masters can do that too." "TURNING MAIN }
\\
\end{tabularx}
\caption{Comparison of language generation with and without conditioning on fMRI brain scans. The context is a snippet from Harry Potter and the anchor words are the Top-5 predictions from our fMRI decoder. The correct word corresponding to the fMRI brain scan is emphasized in boldface.}
\label{tab:gpt2_vs_anchoring}
\end{table}
\renewcommand{\arraystretch}{1}

\subsection{Discussion}

The model presented in this section serves as a proof-of-concept for the application of fMRI decoding to language generation. 
The reason behind our choice of conditioning on the Top-5 decoded words is the following: the model tends to generate text in unpredictable directions, therefore, the four "incorrect" words from the Top-5 can be assimilated to these random directions without overshadowing the effect of the correct anchor. This way, we cover a larger amount of cases while still conditioning towards the correct topic.

Our experiments show that the output of fMRI decoding can guide language generation without loss in fluency. However, to enable real-time fMRI-to-text decoding some improvements are necessary, apart from further improving fMRI-to-word decoding. For instance, in order to account for the delay in fMRI scans due to blood flow, it would be desirable to have a measure of certainty for the decoded word which triggers language generation when the decoder is certain and halts it otherwise. Also it would be necessary to record a new dataset tailored to this application that covers the most important and general concepts a person may want to express, such as "positive", "negative", "happiness", "nature", etc. We are well-aware that there is still a long path before we can reliably turn thoughts into words, for example for coma patients. Nevertheless, we believe that our work provides new tools and ideas to make this possible one day.

\section{Conclusion}

In this work we have presented a model to decode fMRI scans into words that outperforms existing models by a big margin. Furthermore, we have shown that a more realistic task is necessary to understand the performance of decoding models and to this end we propose direct classification. We have run our experiments on the extremely demanding scenario where no data from the target subject is available at training time and demonstrated that our model successfully generalizes to unseen subjects. Based on the results obtained by our decoder, we introduce a strategy for conditioning language generation towards the semantic content of fMRI scans. This way, we contribute towards a real system for translating brain activities to coherent text.

\bibliography{aaai21}
\bibliographystyle{aaai21}

\end{document}